\documentclass[letterpaper, 10pt, conference]{ieeeconf}			

\IEEEoverridecommandlockouts								

\usepackage{graphics}			
\usepackage{epsfig}				
\usepackage{amsmath}			
\usepackage{amssymb}			
\usepackage{subcaption}
\usepackage{diagbox}
\usepackage{algorithm,algpseudocode}

\usepackage[nolist,nohyperlinks]{acronym}

\acrodef{bci}[BCI]{Brain-Computer Interface}
\acrodef{smr}[SMR]{Sensorimotor Rhythm}
\acrodef{eeg}[EEG]{Electroencephalography}
\acrodef{psd}[PSD]{Power Spectral Density}
\acrodef{ssvep}[SSVEP]{Steady State Visually Evoked Potential}
\acrodef{ros}[ROS]{Robot Operating System}
\acrodef{amcl}[AMCL]{Adaptive Monte Carlo Localization}
\acrodef{hrl}[HRL]{Humanoid Robot Localization}

\title{\LARGE \bf
Brain-Computer Interface meets ROS: A robotic approach to mentally drive
telepresence robots*
}

\author{Gloria Beraldo$^{1}$, Morris Antonello$^{1}$, Andrea Cimolato$^{2}$,
Emanuele Menegatti$^{1}$, Luca Tonin$^{3}$%
\thanks{*This research was partially supported by Fondazione Salus Pueri with a
grant from "Progetto Sociale 2016" by Fondazione CARIPARO, by Consorzio Ethics
with a grant for the project "Hybrid Human Machine Interface", and by Omitech
s.r.l. with the grant "O-robot".}%
\thanks{$^{1}$Intelligent Autonomous System Lab, Department of Information
Engineering, University of Padova, Padua, Italy. 
{\tt\small emg@dei.unipd.it} }%
\thanks{$^{2}$Politecnico di Milano in collaboration with Istituto Italiano di
Tecnologia (IIT), Milan, Italy}%
\thanks{$^{3}$Defitech Chair in Brain-Machine Interface (CNBI), \'Ecole
Polytechnique F\'ed\'erale de Lausanne (EPFL), Chemin des Mines 9, CH-1202,
Geneva, Switzerland {\tt\small luca.tonin@epfl.ch}}%
}

\begin{document}

\maketitle
\thispagestyle{empty}
\pagestyle{empty}

\begin{abstract}
This paper shows and evaluates a novel approach to integrate a non-invasive
  \ac{bci} with the \ac{ros} to mentally drive a telepresence robot.
  Controlling a mobile device by using human brain signals might improve the
  quality of life of people suffering from severe physical disabilities or
  elderly people who cannot move anymore. Thus, the \ac{bci} user is able to
  actively interact with relatives and friends located in different rooms thanks
  to a video streaming connection to the robot. To facilitate the control of the
  robot via \ac{bci}, we explore new \ac{ros}-based algorithms for navigation
  and obstacle avoidance, making the system safer and more reliable. In this
  regard, the robot can exploit two maps of the environment, one for
  localization and one for navigation, and both can be used also by the \ac{bci}
  user to watch the position of the robot while it is moving.  As demonstrated
  by the experimental results, the user's cognitive workload is reduced,
  decreasing the number of commands necessary to complete the task and helping
  him/her to keep attention for longer periods of time. 
\end{abstract}

\acresetall
\section{INTRODUCTION}

\ac{bci} technology relies on the real-time detection of specific neural
patterns in order to circumvent the brain's normal output channels of peripheral
nerves and muscles~\cite{Chaudhary2016} and thus, to implement a direct
mind-control of external devices. In this framework, current non-invasive
\ac{bci} technology demonstrated the possibility to enable people suffering from
severe motor disabilities to successfully control a new generation of
neuroprostheses such as telepresence robots, wheelchairs, robotic arms and
software applications~\cite{Birbaumer1999, Galan2008, Leeb2015}. Among the
different \ac{bci} systems developed in the last years, the most promising ones
for driving robotic devices are the so-called endogenous \acp{bci} (e.g., based
on \ac{smr}), where the user autonomously decides when starting the mental task
without any exogenous---visual or auditory---stimulation.

In these systems, neural signals are recorded by non-invasive techniques
(e.g.,~\ac{eeg}) and then, task-related brain-activity is translated into few
commands---usually, discrete---to make the robotic device turn right or left.
Despite the low number of commands provided by non-invasive \acp{bci},
researchers have demonstrated the possibility to drive mobile devices even in
complex situation with the help of a shared control approach~\cite{Leeb2015, Tonin2010, 
Tonin2011}.  The shared control approach~\cite{Goodrich2006} is based
on a seamless human-robot interaction  in order to allow the user to focus his/her
attention on the final destination and to ignore low level problems related to
the navigation task (i.e., obstacle avoidance). The coupling between user's
intention and robot's intelligence allows to contextualize and fuse the
high-level commands coming from the \ac{bci} with the environment information
from the robot's sensors and thus, to provide a reliable and robust
semi-autonomous mentally-driven navigation system. 

In the robotic community \ac{ros}~\cite{Quigley2009} is
becoming the standard \textit{de facto} for controlling different types of
devices. \ac{ros} is a middleware framework that provides a common
infrastructure and several, platform independent, packages (i.e., localization,
mapping, autonomous navigation). Indeed, the most important advantages of
\ac{ros} are its strong modularity and the large and growing community behind:
people can design and implement their own \ac{ros} package with specific
functionalities and thus, distribute it through common repositories.

Although the clear benefits of using \ac{ros}, it is still far to be a standard
adopted in the \ac{bci} community. In \ac{bci} literature, most of the
studies are based on custom and \textit{ad-hoc} implementations of the robotic
part and only few of them clearly reported an integration with common available
tools in \ac{ros}~\cite{Bryan2011, Katyal2014, Arrichiello2017}. The drawback of
this tendency is twofold: on one hand, the lack of standardization makes almost
impossible to check, replicate and validate experimental results. As a matter of
fact in \ac{bci} experiments the technology needs to be tested over a large
population of end-users with severe disabilities and, usually, requires to be
validated by different groups before the acceptance as an effective assistive
tool~\cite{Brunner2015}. On the other hand, home-made control frameworks for
robotic devices imply the adoption of simplified and naive approaches to
fundamental robotic challenges---usually, already solved by the robotic
community---and thus, a limitation of possible applications of the current
\ac{bci} driven neuroprostheses.

This paper aims at showing the benefits of integrating a state-of-the-art
\ac{bci} system and \ac{ros} for controlling a telepresence robot. In
Section~\ref{section:methods}, we describe the \ac{bci} and the robot
adopted as well as our novel navigation algorithm to mentally drive telepresence
robots. In contrast to previous works, it exploits an optimal trajectory
planner and the availability of the environmental map. Furthermore, it is
designed to match the requirements of a semi-autonomous, \ac{bci} driven
telepresence robot. In Section~\ref{section:results}, we evaluate the presented
methods and we showcase the integration with the \ac{bci} system. Finally, in
Section~\ref{section:discussion}, we discuss the results achieved with respect
to similar \ac{bci} based experiments.

\section{METHODS}
\label{section:methods}

\begin{figure*}
\centering
\includegraphics[width=\textwidth]{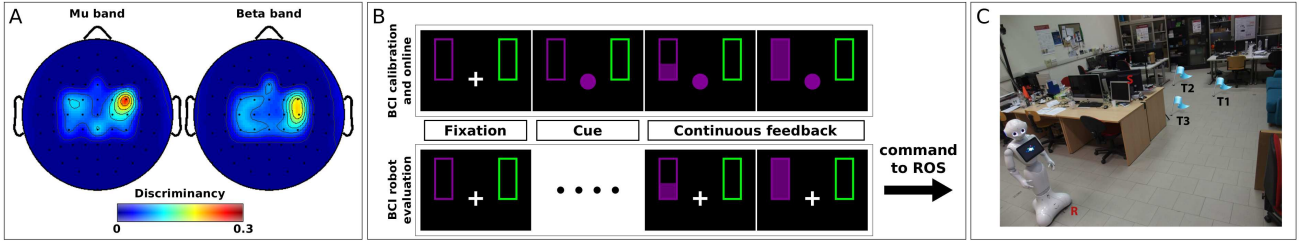}
  \caption{A) Topographic representation of the discriminant features in $\mu$
  and $\beta$ bands used to train the \ac{smr} classifier (fisher score values,
  both hands vs. both feet).  B) Schematic representation of the visual paradigm
  of the \ac{smr} \ac{bci}. Top row: the protocol exploited during the
  calibration and online phases. User is instructed to perform the motor imagery
  task according to a symbolic cue appearing on the screen.  Thus, the \ac{bci}
  classification output is remapped into the movement of the bar. When a bar is
  completely filled, the trial ends. Bottom row: same behavior as before but
  there is no cue and the user decides autonomously which motor imagery task to
  perform to control the robot.  When a bar is completely filled, the related
  command is delivered to the \ac{ros} infrastructure. C) The telepresence robot
  platform (Pepper) and the experimental environment with the three target
  locations.}
\label{fig:methods}
\end{figure*}

\subsection{\acl{bci} system}

In this work, we used a 2-class \ac{bci} based on \acp{smr} to control the
telepresence robot. The user was asked to perform two motor imagery tasks
(imagination of the movement of both hands vs. both feet) to make the robot turn
left or right. Contrariwise to other approaches (e.g., based on evoked
potentials), such a \ac{bci} is based on the decoding of the voluntary
modulation of brain patterns without the need of any external stimulation
repetitively presented to the user.  For this reason, \ac{smr} \acp{bci} have
been widely exploited to successfully drive mobile devices~\cite{Leeb2015, Tonin2010,
Tonin2011, Millan2004, Meng2016}.

The following paragraphs briefly describe the different parts of the \ac{bci}
system developed and used for the study.

\subsubsection{\acs{eeg} acquisition}
A health 24-year-old female and \ac{bci} beginner tried the experiment that was
carried out in accordance with the principles of the Declaration of Helsinki.

\ac{eeg} signals were recorded with an active 16-channel amplifier at 512~Hz
sampling rate, filtered within 0.1 and 100~Hz and notch filtered at
50~Hz (g.USBamp, Guger Technologies, Graz, Austria). Electrodes were placed over
the sensorimotor cortex (Fz, FC3, FC1, FCz, FC2, FC4, C3, C1, Cz, C2, C4, CP3,
CP1, CPz, CP2, CP4) according to the international 10-20 system layout.

\subsubsection{Feature extraction and classification}
\ac{eeg} was pre-processed by applying a Laplacian spatial filter. The \ac{psd}
of the signal was continuously computed via Welch's algorithm (1~second sliding
window, 62.5~ms shift) in the frequency range from 4~to~48~Hz (2~Hz resolution).
Thus, the most discriminative features (channel-frequency pairs,
subject-specific) were extracted and classified online by means of a Gaussian
classifier~\cite{Millan2008} previously trained during the calibration phase
(see Section \ref{section:bciphases}).  Finally, the raw posterior probabilities
were integrated over time to accumulate evidences of the user's intention
according to:

\begin{equation}
  \label{eq:expsmooth}
  p(y_t) = \alpha \cdot p(y_{t-1}) + ( 1 - \alpha ) \cdot  p( y_{t} | x_{t} )
\end{equation}

where $p(y_t | x_t)$ is the probability distribution at time $t$, $p(y_{t-1})$
the previous distribution and $\alpha$ the integration parameter. The
probabilities were showed to the user as a visual feedback
(Fig.~\ref{fig:methods}B). As soon as one of the bar was filled, the command was
delivered to the robot to make it turn right or left.

\subsubsection{Calibration and online phases}
\label{section:bciphases}
As a common practice in \ac{bci} experiments, a calibration phase is required in
order to select the features that each subject can voluntary modulate during
motor imagery tasks and to train the classifier. In this work, the calibration
phase consisted in three runs with 30~trials each where the user was instructed
by symbolic cues about the task to be performed (in total 21~minutes). Thus, we
analyzed the recorded data, we selected the subject-specific features and we
trained the Gaussian classifier. Fig.~\ref{fig:methods}A depicts the spatial and
spectral distributions of the most discriminative features (based on fisher
score values) selected to train the \ac{bci}. The distributions are perfectly
coherent with the brain patterns expected during motor imaginary
tasks~\cite{Chaudhary2016}. 

During the online phase, we evaluated the ability of the \ac{bci} to correctly
detect user's intentions. The user performed three online runs, where he was
asked to control the online \ac{bci} feedback on the screen
(Fig.~\ref{fig:methods}B).

\subsection{Robot}
Our telepresence platform is the Pepper
robot\footnote{https://www.ald.softbankrobotics.com/en/robots/pepper/find-out-more-about-pepper}
by Aldebaran Robotics and SoftBank (Fig.~\ref{fig:methods}C). It is an humanoid
robot designed for human-robot interaction. It features a $1.9$~GHz quad-core
Atom processor and $4$~GB of RAM. It is $1.210$~m high and equipped with an
omnidirectional base of size $0.480 \times 0.425$~m. For obstacle avoidance, it
is provided with two sonars, three bumpers, three laser sensors and three laser
actuators. For vision tasks, the robot has two 2D cameras located in the
forehead, in particular one at the bottom and one at the top, both with a
resolution of $640\times480$~px. For telepresence purposes, we exploited the top
camera to provide a first-person view to the \ac{bci} user by means of the RViz
graphical interface available in \ac{ros}. The Pepper has also an ASUS Xtion 3D
sensor in one of its eyes with a resolution of $320\times240$~px.  However, its
3D data are distorted due to a pair of lenses, positioned in front of it. To
overcome the limitations of the laser and the RGBD-sensor, we built the
environmental maps required for a safe navigation by using data previously
acquired~\cite{Carraro2015} and based on a more powerful Hokuyo URG-04LX-UG01 2D
laser rangefinder able to measure distances from 20~mm to 5.6~m and the more
precise Microsoft Kinect v2. This way, we can still exploit Pepper's sensors for
navigation.

\subsection{\acs{ros}-based Mapping and Localization}

Robot mapping and localization are core functionalities necessary to correctly
navigate in both an autonomous or semi-autonomous way. In particular, we built
the static environmental maps, which are provided to the Pepper for localization
and navigation, with previously acquired data. For the building map process, we
exploited two different methods available in ROS:
\textit{GMapping}~\cite{Grisettiyz2005,Grisetti2007}, and
\textit{OctoMap}~\cite{Hornung2013}. \textit{GMapping} builds a 2D occupancy map
of the environment, while \textit{OctoMap} creates a 3D scene representation,
which can be down-projected to the ground so as to enrich the 2D occupancy map
with higher obstacles visible by the RGBD-sensor but not by the laser. In
particular, the localization module is based on the map built with
\textit{GMapping}, while the navigation one is based on the 2D down-projected
map because of the richer representation of the environment. This way, the
trajectory planner can take into account high obstacles and avoid collisions
(Fig.~\ref{fig:results}A).  As illustrated in Fig.~\ref{fig:results}A, despite
in both maps the planner trajectories seem similar, the path found in the
\textit{Gmapping} based map is less reliable with high obstacles than the one
from \textit{OctoMap}. For instance, in a map built with \textit{GMapping} only
legs of tables are considered, while in the 2D down-projected map from
\textit{OctoMap} they are featured by their flat surfaces.

For localization, the \textit{\ac{amcl}}~\cite{Fox1999} was adopted, with an
adaptive sampling scheme, to make the computation efficient. The
\textit{\ac{hrl}} technique~\cite{Hornung2010}, which is based on the~\ac{amcl}
but uses the 3D \textit{OctoMap}, is also evaluated.

\subsection{ROS-based navigation}
\label{methods:navigation}
Our algorithm allows a semi-autonomous navigation based on a shared control for
\acp{bci}. The main target is twofold: to help the user to successfully drive
the robot and, at the same time, to make him/her feel to have the full control.
Indeed, since the control through an uncertain channel like \ac{bci} can be
complicated, our integration between user and robot is designed so that it
allows the former to only focus on the final destinations; while the latter will
deal with obstacle detection and avoidance, deciding the best trajectory. For
these purposes, we exploited the \ac{ros} \texttt{navigation
stack}\footnote{http://wiki.ros.org/navigation} to localize and move the robot
in the environment according to its sensors, odometry and the static map
provided.

In details, the default behaviour of the robot consists in moving forward and
avoiding obstacles when necessary. The user can control it by his/her brain
activities, delivering voluntary commands (left and right) to turn it to the
corresponding direction. The user's intention is decoded by the \ac{bci} system
and the related command is sent to the \ac{ros} node dealing with navigation
through an UDP packet.

The logic of our algorithm is described in the following pseudo-code.
\begin{algorithm}[h]
\caption{The shared control navigation algorithm}
\begin{algorithmic} [1]
\State $last\_time \leftarrow  \text{current time} $
\While{( $\textsc{IsRobotActive}()$)}
\If{$ \text{current time} - last\_time > CLEAR\_TIME$}
\State $\textsc{ClearCostMap}()$ 
\State $last\_time \leftarrow  \text{current time} $
\EndIf
\If {($\textsc{DefaultBehaviour}()$)}
\State Robot goes forward by a fixed step $S_{x}$
\If{(!Succeeded)}
\State Call $\textsc{RecoveryBehaviour}()$
\EndIf
\Else \Comment BCI command arrival
\State Cancel the current goal
\State Robot turns to the right direction  
\If{(!Succeeded)}
\State Call $\textsc{RecoveryBehaviour}()$
\EndIf
\EndIf
\EndWhile
\Procedure{RecoveryBehaviour}{$ $}
 \State The robot goes back for a fixed time $T$
 \State It turns counter-clockwise by a fixed angle $A$
\EndProcedure
\end{algorithmic}
\label{methods:algorithm}
\end{algorithm}
%
%
At every iteration, our algorithm sends new navigation goals to the robot to
ensure the robot capability of avoiding obstacles in the
environment---especially the dynamic obstacles not represented in the static
map. This way, the planner in the \texttt{navigation stack}, can (re)compute the
best trajectory to reach the target destination even if dynamic obstacles are
presented in the path. In details,  we used the \textit{Dynamic Window
Approach}~\cite{Fox1997} for local planner and  \textit{Dijkstra's algorithm} to
compute the global planner function.  Furthermore, before sending a new goal to
the robot, the corresponding position in the map is checked: the goal is sent to
the robot only if it matches with a free cell in the map, which means that that
cell is not occupied by an obstacle.  Otherwise, the
$\textsc{RecoveryBehaviour()}$ procedure is called to avoid deadlocks by
slightly moving the robot. In detail, the $\textsc{RecoveryBehaviour()}$ makes
the robot go back (if it is possible) and keep turning counter-clockwise until
required. The recovery rotation  takes place always in the same direction (by
fixed angle $A$) to make the robot able to rotate around itself and, thus, to
escape from this undesirable situation. Furthermore, the rotation is carried out
incrementally by sending velocity commands to the robot. If the robot cannot go
back and/or turn due to obstacles, the on-board short-range sonars will stop it.

Even if the target goal corresponds to a free cell in the map, the robot may not
be able to reach it for different technical reasons (e.g., lost connection,
temporary missing frame transformations or unseen obstacles). In this kind of
situations, the procedure $\textsc{RecoveryBehaviour}()$ is called to unstuck
the robot and to ensure a continuous navigation.

Finally, the planner may not still find a valid path due to dynamic obstacles
previously stored in the cost maps but not currently present in the environment.
To avoid such a situation, a clearing operation of the cost maps is done every
$CLEAR\_TIME$.


\subsection{Experimental design}
The experiment was carried out in a typical working space with different
obstacles like tables, chairs, sofa, cupboards, people
(Fig.~\ref{fig:methods}C). The user was seated at position $S$ and the robot
started from position $R$. We defined three target positions $T1$, $T2$, $T3$.
The user was instructed to move the robot from $S$, going through $T1$, $T2$,
$T3$, by only sending mental commands through the \ac{bci}. The default
behaviour of the robot was to move forward and to avoid possible obstacles in
its path. The user perform two repetitions of the task.

\section{RESULTS}
\label{section:results}

\begin{figure*}
\centering
\includegraphics[width=\textwidth]{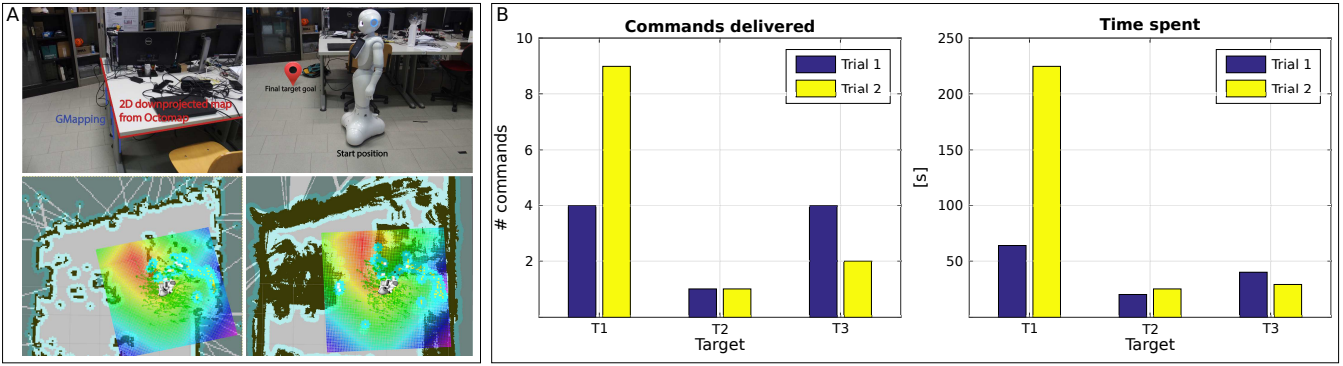}
  \caption{A) Top row: Comparison detection of a desk performed with
  \textit{GMapping} and \textit{2D down-projected map} from \textit{OctoMap}
  (left). Start position and target position for evaluation of robot trajectory
  (right). Bottom row: Best trajectory proposed by the planner when
  \textit{GMapping} (left) or \textit{2D down-projected map} from
  \textit{Octomap} (right) is provided to the robot. B) Number of commands sent
  and time spent by the user  in the two attempts carried out to reach the
  targets $T1$, $T2$, $T3$.}

\label{fig:results}
\end{figure*}

\subsection{Navigation Performances}
In this work, we considered different combinations of the two kind of maps
provided as input to the robot and the method used for localization. More
precisely, we examined the performance in terms of number of delocalizations and
collisions against obstacles in the experimental environment by simulating 150
random commands delivered by the \ac{bci} system. Table~\ref{results:maps}
depicts the results achieved.

The combination between the \textit{2D down-projected map} from \textit{Octomap}
and \textit{GMapping}  together with the \ac{amcl} localization method
represents a good compromise (3 number of delocalizations and 2 collisions)
between providing a more detailed map to the robot and using a 2D and fast
computation localization algorithm (Table~\ref{results:maps}, second row).
Indeed, although the third approach (with \ac{hrl} localization method) allowed
the lowest number of possible collisions, it requires higher computational power.
This is due not only to the 3D \textit{OctoMap} used for navigation, but also to
the fact that \ac{hrl} does not exploit an adaptive sampling scheme adjusting
the number of samples. 

\begin{table}[]
\centering
\resizebox{0.48\textwidth}{!}{
\begin{tabular}{|c|c|c|c|c|}
\hline
\textbf{\begin{tabular}[c]{@{}c@{}}Navigation \\ Map\end{tabular}}
  & \textbf{\begin{tabular}[c]{@{}c@{}}Localization \\ Map\end{tabular}} &
	\textbf{\begin{tabular}[c]{@{}c@{}}Localization \\ Method\end{tabular}} &
	  \textbf{\begin{tabular}[c]{@{}c@{}}Number of \\
		delocalizations\end{tabular}} &
		\textbf{\begin{tabular}[c]{@{}c@{}}Number of \\ collisions\end{tabular}}
		  \\ \hline
\textit{GMapping} & \textit{GMapping}
  & \ac{amcl} & 6 & 5
  \\ \hline
\textit{\begin{tabular}[c]{@{}c@{}}2D down-projected \\ map from
OctoMap\end{tabular}} & \textit{GMapping} & \ac{amcl}
  & 3 & 2
  \\
  \hline
\textit{\begin{tabular}[c]{@{}c@{}}2D down-projected \\ map from
OctoMap\end{tabular}} & 3D \textit{Octomap} & \ac{hrl}
  & 10 & 1
  \\
  \hline
\end{tabular}}
\caption{Evaluation of different combinations of the two kind of maps provided
  as input to the robot and the method used for localization. Data were
  acquired by simulating 150 random commands delivered by the \ac{bci} system
  for each approach.}
\label{results:maps}
\end{table}

\subsection{\acs{bci} driven telepresence}
For the integration of \ac{bci} and \ac{ros}, we analyzed the number of
\ac{bci} commands delivered by the user to reach the targets $T1$, $T2$, $T3$
and the corresponding times (Fig. \ref{fig:results}B). In average, the user
delivered 3.0$\pm$1.3 commands and employed 34.5$\pm$32.2~s (median and standard
error) to reach each target. The number of commands and the time required were
low for all targets in each repetitions (except for $T1$, second repetition,
where the user sent few wrong \ac{bci} commands to the robot). 

Furthermore, we evaluates the importance of shared control by comparing the
\ac{bci} with a manual control. In this case, we asked the user to repeat the
experiment controlling the Pepper robot with discrete commands sent by the
keyboard but without the assistance of the shared control for obstacle
avoidance. The ratio between the number of commands in the two modalities
(\ac{bci} with shared control and the manual without shared control) was 80.9\%
and the ratio between times was 114.5\%.

It is possible to notice that the number of commands increased in the manual
modality.  This means that without shared control, the \ac{bci} user has to send
more commands to the robot, increasing the necessary cognitive workload.
Especially, in that case in which the robot is blocked because in its
neighborhood there are some obstacles that make it stuck. However, the time
spent is less using the keyboard, due to the time required by the \ac{bci}
system before delivering a commands to the robot.

\section{DISCUSSION}
\label{section:discussion}
The main objective of this study was to demonstrate for the first time the
potentialities and the perspectives related to the integration between \ac{ros}
and a \ac{bci} system.  Modularity of \ac{ros} allows robotic community to
exchange and distribute repositories besides the platform adopted
\cite{Quigley2009}.  This particular aspects is what makes \ac{ros} very
appealing in assistive robotics.  \ac{ros} is able to provide a common
infrastructure where developers can either decide to share their novel
approaches or adopt external tools through use of common repositories. In
this context, this work aimed at promoting collaboration of multiple disciplines
in order to design a semi-autonomous \ac{eeg} driven navigation for telepresence
robots. Moreover, integration of the \ac{bci} with \ac{ros} allowed testing the
system on Pepper robotic platform, never experienced in \ac{bci} driven
teleoperation.  Second purpose of this work was the development of a novel
approach for assistive robotic navigation based on multiple-maps input under
\ac{bci} shared control.  Our approach demonstrated the possibility to make
obstacle avoidance more reliable and, therefore, the navigation safer.

Results comparison between other \ac{bci} studies may be complicated and
not always meaningful, due to different testing conditions. However, it is worth
to notice that with respect to previous works, results are consistent in terms
of ratio between \ac{bci} and manual input both for time intervals and number of
commands. Our work reported a 114.5\% ratio for time intervals between the two
modalities, in line to~\cite{Leeb2015} where it was estimated 109$\pm$11\% for
the end-users and 115$\pm$10\% for healthy ones. Similar trend results are
reported in~\cite{Tonin2011} where both type of users in average achieved
118.5$\pm$19\% ratio between the two modalities. Furthermore, the 19.1\%
decreasing in number of commands recorded in our experiment was in perfect
agreement with~\cite{Leeb2015} and~\cite{Tonin2011} where similar reduction was
reported.  These preliminary results suggest the possibilities and the
advantages of using \ac{ros} in \ac{bci} driven telepresence applications.

The proposed \ac{bci} system is one of the few working on the top of a \ac{ros}
framework~\cite{Bryan2011, Katyal2014, Arrichiello2017} and, among them, the
only one supporting an endogenous \ac{smr} based \ac{bci}. As in the case of
previous works~\cite{Leeb2015}, the designed semi-autonomous control reduced the
user's fatigue (in terms of number of commands required to reach the target).
Furthermore, \ac{ros} was fundamental in our approach to enable communication
among different software and hardware modules and it was essential to overcome
\ac{bci} limitations by exploiting well-established robotic solutions for
obstacle avoidance and navigation.

Modern \ac{bci} teleoperating systems are not mature enough to be exploited in
the daily life despite the promising results. This divergence is due mainly to
different complexities between testing conditions and home-like environments.
High density of obstacles and non-uniform space distribution make impracticable
to use mentally driven systems in such situations. In fact, platform control
could result stressful and exhausting for the end-user, even with obstacle
avoidance assistance. In order to provide relief to the user in such conditions,
our proposal was to include a localization algorithm in navigation. Direct
interaction with this module output conveyed better understanding of robotic
platform state and allowed the user to plan in advance the navigation, dealing
with delays in command delivery. Maps localization, moreover, was designed to
admit path planning strategies in the obstacle avoidance algorithm, promoting an
evolution of the shared control approach. Previous implementations of the shared
control were able to detect the obstacle and modify the trajectory in order to
avoid collisions~\cite{Leeb2015, Tonin2010, Tonin2011}. However, since algorithm was nor
provided with an intermediate goal nor with favorite direction, once the hurdle
was evaded, it was user's burden to put the teleoperated device back on track.
Contrariwise, our novel implementation permits to identify an obstacle
and plan accordingly a new trajectory in order to avoid collision but, at the
same time, not deviating from the direction imposed from the user. Fundamental
for navigation in hostile areas, involving for example moving obstacle, was the
\textit{recovery procedure} (Section~\ref{methods:navigation}). This feature,
combined with the path planner in partial target computation, avoided algorithm
to fail in case of conflicts in the occupancy map.

Future directions of the proposed work will be to first improve Pepper 3D
vision, that have been the main limitation of the platform. Extrinsic and
intrinsic calibration of RGB-D cameras and the related point cloud noise
reduction could be resolved using \textit{RGB-D Calibration packet} proposed
by~\cite{Basso2014}. This improvement will allow 3D point cloud localization
integration, augmenting its reliability in obstacle avoidance and adaptability
to complex environment.  With regard to navigation, it is intention of the
authors testing as input a full \textit{3D OctoMap},  either for localization
and trajectory estimation toward partial target. This should generate a similar
navigation control to \texttt{3d\_navigation stack} \ac{ros} package, not
available for recent versions of the robotic operating system.  Final
improvements will be addressed to reduce workload on the user. First approach to
pursuit will be to integrate classification in object recognition, correlating
class of obstacles to actions to take. People detection and tracking could
represent additional features to relieve users from the burden of navigation and
control attention.  Additional relaxation in user control can be provided using
Intentional Non-Control~\cite{Tonin2017}, which should detect when the user does
not want to perform any motor task. Algorithm therefore should act in slowing or
even stopping integration of the command delivery output, when such condition is
present.

The benefits of integrating \ac{ros} in \ac{bci} driven devices are not limited
to telepresence purposes. For instance, the robustness and the reliability
provided by \ac{ros} can be exploited to encourage the use of the \ac{bci} in
more sensitive domains such as car applications~\cite{yu2016},
rehabilitation~\cite{daly2015}, pediatric interventions~\cite{Breshears2011} and
pain mitigation~\cite{yoshida2016}.

\addtolength{\textheight}{-12cm}   



\section*{ACKNOWLEDGMENT}
This research was partially supported by Fondazione Salus Pueri with a grant
from "Progetto Sociale 2016" by Fondazione CARIPARO, by Consorzio Ethics with a
grant for the project "Hybrid Human Machine Interface", and by Omitech srl with
the grant "O-robot".  We thank Dott. Roberto Mancin and Omitech s.r.l. for
hardware support.

\bibliographystyle{IEEEtran}
\bibliography{IEEEabrv,root}

\end{document}